\documentclass[letterpaper, 10 pt, conference]{ieeeconf}  

\IEEEoverridecommandlockouts                              

\usepackage{epsfig} 
\usepackage{amsmath} 
\usepackage{amssymb}  
\usepackage{booktabs}
\usepackage{array}
\usepackage{float}
\usepackage{multirow}
\usepackage{subfigure}
\usepackage{url}

\title{\LARGE \bf
Generic Vehicle Tracking Framework Capable of Handling Occlusions Based on Modified Mixture Particle Filter 
}

\author{Jiachen Li,
	Wei Zhan 
	and Masayoshi Tomizuka
\thanks{J. Li, W. Zhan, and M. Tomizuka are with the Department of Mechanical Engineering, 
        University of California, Berkeley, CA 94720, USA
        (e-mail: {\tt\small jiachen\_li, wzhan, tomizuka@berkeley.edu})}
}

\begin{document}
	
\maketitle
\thispagestyle{empty}
\pagestyle{empty}
\begin{abstract}
Accurate and robust tracking of surrounding road participants plays an important role in autonomous driving. 
However, there is usually no prior knowledge of the number of tracking targets due to object emergence, object disappearance and false alarms.
To overcome this challenge, we propose a generic vehicle tracking framework based on modified mixture particle filter, which can make the number of tracking targets adaptive to real-time observations and track all the vehicles within sensor range simultaneously in a uniform architecture without explicit data association. Each object corresponds to a mixture component whose distribution is non-parametric and approximated by particle hypotheses. 
Most tracking approaches employ vehicle kinematic models as the prediction model. However, it is hard for these models to make proper predictions when sensor measurements are lost or become low quality due to partial or complete occlusions. Moreover, these models are incapable of forecasting sudden maneuvers.
To address these problems, we propose to incorporate learning-based behavioral models instead of pure vehicle kinematic models to realize prediction in the prior update of recursive Bayesian state estimation.
Two typical driving scenarios including lane keeping and lane change are demonstrated to verify the effectiveness and accuracy of the proposed framework as well as the advantages of employing learning-based models.
\end{abstract}

\section{INTRODUCTION}
Accurate and efficient vehicle tracking plays an significant role in autonomous driving since it is the basis of state estimation for surrounding vehicles in real time which is a prerequisite for proper planning and control \cite{r1}. However, there remains three principal challenges for multi-target tracking: 1) the number of tracking targets is usually unknown and may fluctuate over time due to new object emergence, existing object disappearance and false alarms; 2) the tracking targets may be partially or even completely occluded by surrounding objects which leads to low-quality or missing measurements; 3) It is hard to track highly dynamic driving maneuvers such as sudden acceleration or deceleration.
In order to overcome these challenges, a robust and accurate tracking framework is required which can 
1) adaptively adjust the number of objects; 
2) provide relatively accurate prediction during occlusion periods; and 
3) keep track of immediate state changes.

In recent years, a large number of studies on multi-target tracking have been put forward which basically fall into two categories.
The first category attempts to employ deep neural networks combined with computer vision techniques to realize End-to-End tracking by detection on images and videos \cite{r2}. 
However, a huge amount of data is required to train the detection network which demands an enormous data base and high computation support. 
The second category aims at estimating the probability distribution of tracking target state via various probabilistic inference methods (e.g. Bayesian state estimation) with sequential observations. 
Among them, Kalman filter (KF) is proved to be the optimal estimator for linear systems with Gaussian-distributed states \cite{r3}\cite{r4} and Extended Kalman filter (EKF) is utilized for non-linear systems \cite{r5}. 
However, since it is hard to accurately represent the state distribution by simple multivariate Gaussian distributions in the real world, Particle filter (PF) has advantages over variants of Kalman filter since it has no limitations on the form of the system and state distributions \cite{r6}-\cite{r11}. 
In \cite{r12} and \cite{r14}, grid-based particle filters were used to estimate the dynamics of the traffic environment. 
However, explicit data association is required in above methods, which has significant effects on tracking results.
Moreover, the algorithm complexity and computation cost increase exponentially as the number of objects grows \cite{r1}.

Therefore, we propose a uniform tracking framework based on modified mixture particle filter (MPF) \cite{r15} to track multiple objects simultaneously without explicit data association so that the number of tracking targets and particles can be set adaptive to observations. 
Also, the mixture representation is more effective at capturing multiple modes than a single distribution. 
To the best of our knowledge, the concept of mixture tracking was first proposed in \cite{r15} to maintain multi-modality and tested on visual tracking of football players. 
In \cite{r16}, Phi-Vu et al. utilized the same technique on motorcycle visual tracking which achieved encouraging accuracy. However, neither did they provide a uniform tracking framework for multiple objects nor did they handle occlusion problems. 

Vehicle kinematic models are widely used as the state transition model at the prediction step in recursive state estimation. 
The simplest linear models are the Constant Velocity Model (CVM) and Constant Acceleration Model (CAM). 
More complicated models, such as Constant Steering Angle and Velocity Model (CSAVM) and Constant Steering Angle and Acceleration Model (CSAAM), take into account the correlation between the velocity and the yaw rate, which are also known as bicycle models \cite{r17}.
Using these models can achieve encouraging tracking performance when sensor measurements are of high-quality. 
However, they are incapable of making long-term predictions when sensor measurements are lost or with low quality for a relatively long period. 
To deal with the problem, we propose to enhance the capability of prediction models in addition to improving the detection algorithms. 

Many research efforts have been devoted to design more sophisticated driver models considering the impacts from surrounding vehicles \cite{r18}-\cite{r24}. 
The Intelligent Driver Model (IDM) is a representative car-following model which describes the dynamics of the position and velocity of a single vehicle \cite{r21}. In \cite{r22}, Stefan et al. proposed to use IDM and particle filtering to make probabilistic long-term prediction for car-following behavior in highway scenarios. 
\cite{r23} brought forward a Gaussian mixture model based prediction method to illustrate the benefits of the data driven approach for the longitudinal prediction of vehicles.
In \cite{r24}, Constrained Policy Net was proposed to achieve safe and feasible motion planning and prediction in urban scenarios.

Also, a lot of studies have focused on dealing with visual tracking under occlusions. For example, \cite{r6} presented a particle filtering approach to handle vehicle tracking under partial and complete occlusion for traffic video surveillance systems. 
However, only a few investigations devoted to handle occlusions in kinematic state tracking.
\cite{r25} proposed to use a dynamics model accounting for driving behaviors of road participants and a hybrid Gaussian mixture model (hGMM) to obtain multiple hypotheses where the standard object trackers are augmented by discrete states. However, the tracking method could only track an individual vehicle and explicit data association was needed. Moreover, they did not take into account the potential effects of surrounding vehicles on the tracked vehicle, which may be insufficient for reasonable prediction.

In this work, we take advantage of a learning-based behavioral model in the proposed tracking framework as the system dynamics model for its capability of interacting with surrounding vehicles as well as learning to predict feasible and reasonable motions from real-world driving data. This approach has great advantages on tracking sudden maneuvers as well as reducing tracking variance.

The remainder of the paper is organized as follows. Section II presents the generic vehicle tracking framework based on modified mixture particle filter. Section III provides the details of learning-based prediction model. In Section IV, two case studies including lane keeping and lane change scenarios are illustrated. Finally, conclusions are drawn in Section V followed by the future work.

\section{Generic Vehicle Tracking Framework}
In this section, we first introduce the theoretical basis of the mixture tracking which is a combination of recursive Bayesian state estimation and mixture model representation \cite{r15}. The formulation and mechanisms of the modified mixture particle filter are then illustrated. At last, the generic vehicle tracking framework is demonstrated.  

\subsection{Theoretical Basis}
The theoretical basis of the proposed framework is recursive state estimation with sequential observations. We denote the state of a tracking target at time step $k$ as $\mathbf{x}_k$, the augmented state at time $k$ as $\mathbf{e}_k$ which includes intermediate variables and exterior information, and the observations up to time step $k$ as $\mathbf{z}^{k} = (\mathbf{z}_1 \cdots \mathbf{z}_k)$. The Bayesian state estimation in this work contains two steps:\\
\textbf{Prior Update}: 
\begin{equation}
\begin{split}
f(\mathbf{x}_k|\mathbf{z}^{k-1}) = 
\iint f(\mathbf{x}_k|\mathbf{x}_{k-1},\mathbf{e}_{k-1})
f(\text{d}\mathbf{x}_{k-1},\text{d}\mathbf{e}_{k-1}|\mathbf{z}^{k-1})
\end{split}
\end{equation}
\textbf{Measurement Update}:
\begin{equation}
f(\mathbf{x}_k|\mathbf{z}^{k}) = \frac{f(\mathbf{z}_k|\mathbf{x}_k)f(\mathbf{x}_k|\mathbf{z}^{k-1})}{\int f(\mathbf{z}_k|\mathbf{x}_k)f(\text{d}\mathbf{x}_k|\mathbf{z}^{k-1})}
\end{equation}
where $f(\cdot)$ represents the probability density function.
A system dynamics model and a measurement model are required to obtain the state transition distribution and measurement likelihood. The recursion is initialized with a known distribution $f(\mathbf{x}_0|\mathbf{z}_0)$ according to the initial observation.

The mixture tracking formulation is a non-parametric representation of prior and posterior multi-modal distributions which can be recursively updated similar to canonical Bayesian state estimation. The posterior state distribution can be represented as a mixture model:
\begin{equation}
f(\mathbf{x}_k|\mathbf{z}^{k}) = \sum_{m=1}^{M} \pi_{m,k}f_{m}(\mathbf{x}_k|\mathbf{z}^{k})
\end{equation}
where $M$ is the number of mixture components, $\pi_{m,k}$ is the mixture weight for the $m$-th component at time step $k$ and $\sum_{m=1}^{M} \pi_{m,k}=1$. Assuming that the posterior state distribution at time step $k-1$, i.e. $f(\mathbf{x}_{k-1}|\mathbf{z}^{k-1})$ has been obtained from the last measurement update, we can calculate the new prior state distribution straightforwardly by
\begin{equation}
\begin{split}
f(\mathbf{x}_k|\mathbf{z}^{k-1}) = &\sum_{m=1}^{M} \pi_{m,k-1} \iint [f_m(\mathbf{x}_k|\mathbf{x}_{k-1},\mathbf{e}_{k-1})  \\ 
&\times f_{m}(\text{d}\mathbf{x}_{k-1},\text{d}\mathbf{e}_{k-1}|\mathbf{z}^{k-1})].
\end{split}
\end{equation}
When a new measurement is obtained, the prior state distribution is substituted into (2), which leads to 
\begin{equation}
\begin{split}
f(\mathbf{x}_k|\mathbf{z}^{k}) = 
\frac{\sum_{m=1}^{M} \pi_{m,k-1} f_m(\mathbf{z}_k|\mathbf{x}_k) f_{m}(\mathbf{x}_k|\mathbf{z}^{k-1})}{\sum_{n=1}^{M} \pi_{n,k-1} \int f_n(\mathbf{z}_k|\mathbf{x}_k) f_{n}(\text{d}\mathbf{x}_k|\mathbf{z}^{k-1})}.
\end{split}
\end{equation}
The new posterior distribution and mixture weight for the $m$-th component can be obtained through following equations:
\begin{equation}
f_{m}(\mathbf{x}_k|\mathbf{z}^{k}) = \frac{f_m(\mathbf{z}_k|\mathbf{x}_k) f_{m}(\mathbf{x}_k|\mathbf{z}^{k-1})}{\int
f_m(\mathbf{z}_k|\mathbf{x}_k) f_{m}(\text{d}\mathbf{x}_k|\mathbf{z}^{k-1})},
\end{equation}
\begin{equation}
\pi_{m,k} = \frac{\pi_{m,k-1} f_{m}(\mathbf{z}_k|\mathbf{z}^{k-1})}{\sum_{n=1}^{M}\pi_{n,k-1} f_{n}(\mathbf{z}_k|\mathbf{z}^{k-1})}.
\end{equation}
The above recursion can be applied to each individual component and the components only interact through the mixture weights.

\subsection{Modified Mixture Particle Filter}
To approximate the mixture tracking recursion, we propose a modified mixture particle filter formulation. The particle state contains five sets of variables: vehicle state $\mathbf{x}$, particle weight $w$, particle raw weight $\tilde{w}$, component index $c$ (the index of the component that the particle belongs to), and component weight $\pi$.
Note that each particle in this formulation is self-contained.
The particle $i$ at time step $k$ is denoted as $\mathbf{p}^{(i)}_{k} = [\mathbf{x}^{(i)}_k \ w^{(i)}_{k} \ \tilde{w}^{(i)}_{k} \ c^{(i)}_{k} \ \pi^{(i)}_{k}]^{T}$.
The mixture posterior distribution can be approximated with a great number of such particles:
\begin{equation}
\hat{f}(\mathbf{x}_{k}|\mathbf{z}^{k}) = \sum_{m=1}^{M} \pi_{m,k} \sum_{i \in \mathcal{C}_{m}} w^{(i)}_{k} \delta ({\mathbf{x}^{(i)}_{k}} - \mathbf{x}_{k})
\end{equation}
where $\delta (\cdot)$ represents Dirac delta function, $\sum_{m=1}^{M} \pi_{m,k}=1$, $\sum_{i \in \mathcal{C}_{m}} w^{(i)}_{k}=1, m=1...M$ and $\mathcal{C}_{m}$ denotes the set of indices of the particles belonging to the $m$-th component. Since the mixture components are updated independently in the mixture tracking algorithm, in the same way, the particles within each component can also be updated independently and the particles in different components interact only through their corresponding component weights. For instance, the particle set at time step $k$ for the $m$-th component $\{\mathbf{p}^{(i)}_{k}, i \in \mathcal{C}_{m}\}$ is properly sampled from distribution $f_{m} (\mathbf{x}_{k-1}|\mathbf{z}^{k-1})$ and then updated according to prior update and measurement update. The new particles are re-weighted by
\begin{equation}
w^{(i)}_{k} = \frac{\tilde{w}^{(i)}_{k}}{\sum_{j \in \mathcal{C}_{m}} \tilde{w}^{(j)}_{k}},
\end{equation}
\begin{equation}
\tilde{w}^{(i)}_{k} = \frac{w^{(i)}_{k-1} f_m(\mathbf{z}_k|\mathbf{x}^{(i)}_{k})f_m(\mathbf{x}^{(i)}_{k}|\mathbf{x}^{(i)}_{k-1})}{f_m(\mathbf{x}^{(i)}_{k}|\mathbf{x}^{(i)}_{k-1}, \mathbf{z}_{k})}.
\end{equation}
Then the new particle set $\{\mathbf{p}^{(i)}_{k}, i \in \mathcal{C}_{m}\}$ can properly approximate the posterior distribution at time step $k$. Finally, the component weights can be updated as
\begin{equation}
\pi_{m,k} = \frac{\pi_{m,k-1} \tilde{w}_{m,k}}{\sum_{n=1}^{M}\pi_{n,k-1} \tilde{w}_{n,k}}, \ \tilde{w}_{m,k} = \sum_{i \in \mathcal{C}_{m}} {\tilde{w}^{(i)}_{k}}.
\end{equation}
It is necessary to resample the particles for each mixture component like standard particle filters to avoid weight degeneracy.

\subsection{Generic Vehicle Tracking Framework}
After constructing the modified mixture particle filter (MMPF), we can employ it in the tracking framework which is a closed loop consisting of three stages: initialization, particle update and mixture update (see Fig. 1). The details of each step are presented below.
\begin{figure}[htbp]
	\centering
	\epsfig{figure=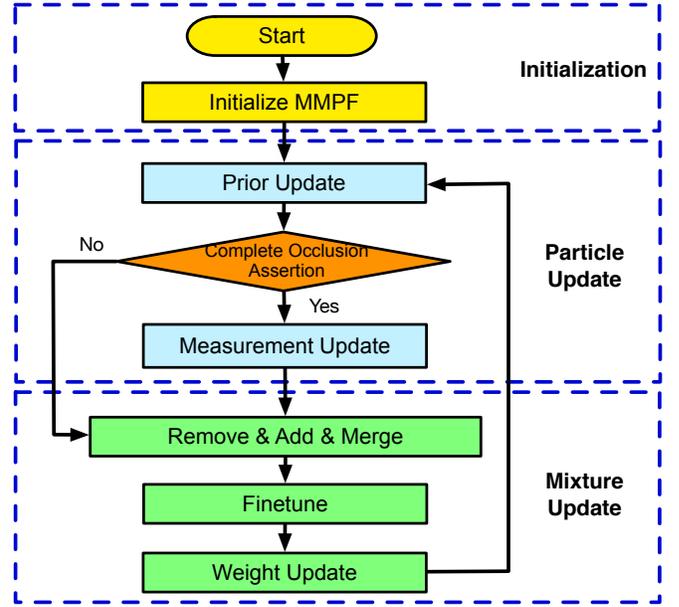,width=\columnwidth,height=8cm}
	\caption{Flow Diagram of Generic Vehicle Tracking Framework}
\end{figure}
\subsubsection{Initialization} The initial particles are drawn around the detected vehicles in the first frame. 
\subsubsection{Prior update} A well-trained learning-based driver behavioral model is utilized as the dynamics model to propagate the particles. There is no restriction on the model architecture as long as it takes as input the state information and/or exterior information at current time step and outputs the state hypotheses for future time steps. Each particle propagates independently.
\subsubsection{Complete occlusion assertion} If a tracking target is asserted to be completely occluded by the detection algorithm, then the corresponding particles do not perform measurement update. 
\subsubsection{Measurement update} We set the measurement likelihood of each particle as the largest likelihood value with respect to all measurements. Then the measurement update is applied and the particles are resampled according to the new weights. Note that there is no explicit data association between observations and mixture components in this step.
\subsubsection{Remove \& Add \& Merge} The components are removed if they leave the observation area or their weights are less than a threshold $r$. 
If the number of particles assigned to a certain measurement in the last step is less than a threshold $a$, we treat the measurement as emergence of a new object. Thus a new component will be added and new particles are drawn.
After that, the components with significant overlap are merged.
\subsubsection{Finetune} Since the component number may be modified in the last step, it is necessary to finetune the particle representation. We use the $k$-medoids method \cite{r24-1} to recluster the particles where $k$ is set to be the new component number, which does not change the approximated posterior distribution. The particles may transfer among different mixture components after reclustering.
\subsubsection{Weight update} The particle and component weights are re-computed and normalized for newly clustered particles.

\section{Behavioral Model}
In this section, we introduce our learning-based behavioral model for dealing with the motion predictions under occlusions. We assume that for partially occluded vehicles, very noisy measurements can be obtained from the detection algorithm; while for completely occluded ones, we have no information about their motions and intentions until they appear in our visible region again. 
In this paper, without loss of generality we use a Gaussian mixture model (GMM) as the prediction model.

\subsection{Gaussian Mixture Distribution Fitting}
The Gaussian mixture distribution can be written as a linear superposition of multiple Gaussians with the form 
\begin{equation}
f(\mathbf{\zeta}) = \sum_{g=1}^{N} \pi_{g} \mathcal{N}(\mathbf{\zeta}|\mathbf{\mu}_{g}, \mathbf{\Sigma}_{g})
\end{equation}
where $\sum_{g=1}^{N} \pi_{g} = 1$, $\mu_{g}$ and $\Sigma_{g}$ are the mean and covariance of the $g$-th Gaussian distribution, and $\mathbf{\zeta}$ is the training dataset. In each training sample, the input and output are stacked into a column vector which is denoted as $\zeta = [\ \mathcal{I} \ | \ \mathcal{O} \ ]^{T}$,
where $\mathcal{I}$ signifies input and $\mathcal{O}$ signifies output. The dimensions of input and output are arbitrary.

The Gaussian mixture distribution is fitted to training data by Expectation-Maximization (EM) algorithm. The component number is decided by the Bayesian information criterion (BIC). The initial means and covariances are set to be the results of $k$-means clustering algorithm and initial component weights can be set as the fractions of data points assigned to the corresponding cluster. We can obtain the estimated joint distribution after the convergence of the  log-likelihood function which is given by
\begin{equation}
\ln f(\mathbf{\zeta}|\pi,\mathbf{\mu}, \mathbf{\Sigma}) = \sum_{n=1}^{N_{\zeta}} \ln \{\sum_{g=1}^{N} \pi_{g} \mathcal{N}(\mathbf{\zeta}|\mathbf{\mu}_{g}, \mathbf{\Sigma}_{g})\}
\end{equation}
where $N_{\zeta}$ is the number of training samples.

\subsection{Prediction Method}
After the GMM has been well fitted to the training dataset, the goal of the prediction method is to obtain the conditional distribution of output $\mathcal{O}$ given an input $\mathcal{I}$, i.e. $f(\mathcal{O}|\mathcal{I})$. Since the Gaussians in a GMM are independent and just interact through mixture component weights, the whole conditional distribution can be calculated as a linear combination of conditional distributions of each Gaussian. Considering the $g$-th component, the mean and covariance matrix can be decomposed according to input and output dimensions:
\begin{equation}
\mu_{g} = [\ \mu_{g,\mathcal{I}} \ \vline \ \mu_{g,\mathcal{O}} \ ]^{T},
\end{equation}
\begin{equation}
\Sigma_{g} = \left[ \ \begin{array}{c|c}
\Sigma_{g,\mathcal{I}} & \Sigma_{g,\mathcal{I},\mathcal{O}} \\
\hline
\Sigma_{g,\mathcal{O},\mathcal{I}} & \Sigma_{g,\mathcal{O}} \\
\end{array} \ \right].
\end{equation}
The conditional output mean and covariance and the corresponding component weight can be obtained by the following equations:
\begin{equation}
\mu_{g,\mathcal{O}|\mathcal{I}} = \mu_{g,\mathcal{O}} + \Sigma_{g,\mathcal{O},\mathcal{I}} \Sigma^{-1}_{g,\mathcal{I}} (\mathcal{I}-\mu_{g,\mathcal{I}}),
\end{equation}
\begin{equation}
\Sigma_{g,\mathcal{O}|\mathcal{I}} = \Sigma_{g,\mathcal{O}} - \Sigma_{g,\mathcal{O},\mathcal{I}} \Sigma^{-1}_{g,\mathcal{I}} \Sigma_{g,\mathcal{I},\mathcal{O}},
\end{equation}
\begin{equation}
\pi_{g|\mathcal{I}} = \frac{\pi_{g} f(\mathcal{I}|\mathcal{N}(\mu_{g}, \Sigma_{g}))}{\sum_{n=1}^{N} \pi_{n} f(\mathcal{I}|\mathcal{N}(\mu_{g}, \Sigma_{g}))}.
\end{equation}
At this stage, the conditional distribution of output $\mathcal{O}$ given a certain input $\mathcal{I}$ is obtained and utilized to make short-term or long-term predictions.
While a bunch of GMM-based approaches try to make deterministic predictions by calculating the weighted mean to find the most probable state, ours attempts to sample particle hypotheses from the multi-modal distribution  to incorporate uncertainties.

\section{Case Study} 
In this section, two typical behaviors in highway scenarios are investigated to validate the vehicle tracking framework and the proposed prediction model: lane keeping and lane change. The data source, experiment details are presented and results are analyzed.

\subsection{Data Source}
The training, validation and test data for these two cases were extracted from the Next Generation Simulation (NGSIM) dataset which is available online \cite{r26}.
The vehicle trajectory data was collected on southbound US101 in Los Angeles, CA, within an area approximately 640 meters (2,100 feet) in length and consisted of five mainline lanes. The dataset provides location, velocity and acceleration information of each vehicle every 0.1 second, which is suitable for training the behavioral model.
\begin{figure*}[!tbp]\label{partial_occ}
	\centering
	\epsfig{figure=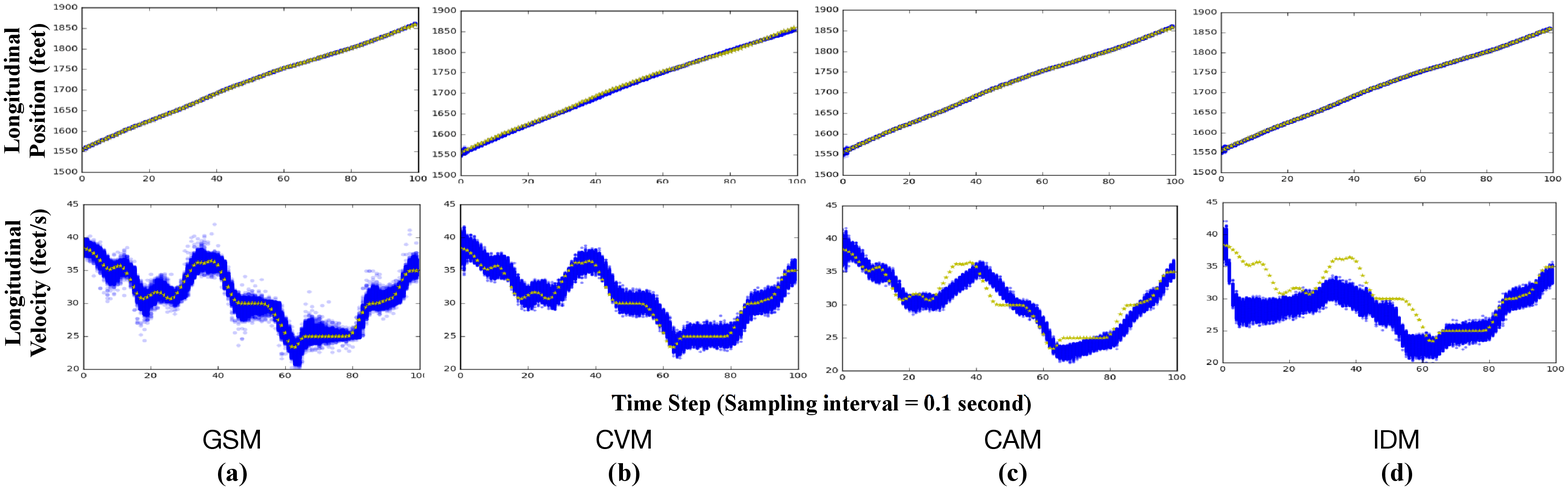,width=\textwidth,height=4.8cm}
	\caption{Longitudinal Position and Velocity Tracking Results Comparison for Partial Occlusion (Lane Keeping)}
\end{figure*}
\begin{figure*}[!tbp]\label{total_occ}
	\centering
	\epsfig{figure=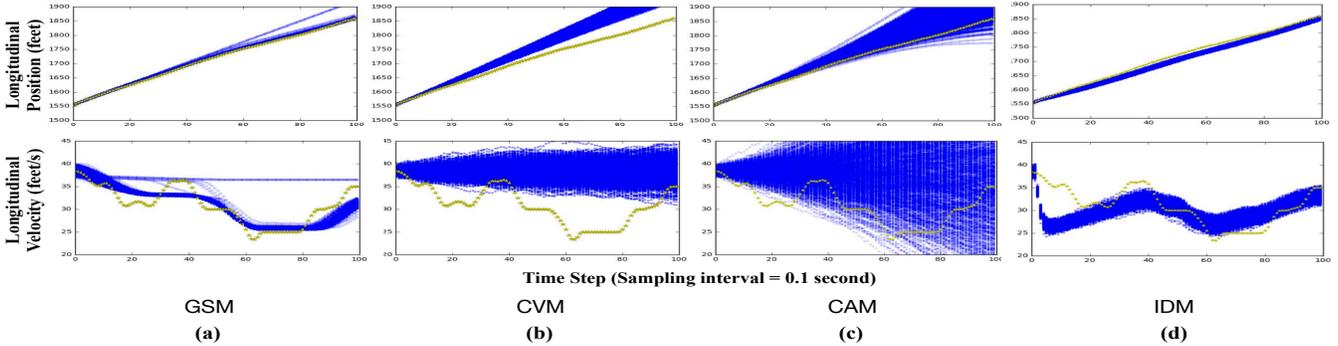,width=\textwidth,height=4.8cm}
	\caption{Longitudinal Position and Velocity Tracking Results Comparison for Complete Occlusion (Lane Keeping)}
\end{figure*}
\subsection{Case 1: Lane Keeping}
In this case, each vehicle is assumed to perform car-following behavior without changing lanes. Although each vehicle in the observation area needs to be predicted for tracking purpose, we choose one of them as object of study without loss of generality.
The vehicle state contains lateral position, longitudinal position, velocity and acceleration.
A Gaussian mixture model (GMM) for short-term prediction was trained to forecast the vehicle state only one time step forward (0.1 second) which is named as GMM Short-term Model (GSM). 

We denote the $i$-th training sample as 
\begin{equation}{\label{short_train2}}
\mathcal{I}_{i} = [ \ x^{fol}_{k} \  d^{rel}_{k} \  v^{fol}_{y,k} \  a^{fol}_{y,k}  \ v^{lead}_{y,k} \ ]
\end{equation}
\begin{equation}{\label{short_train3}}
\mathcal{O}_{i} = [ \ x^{fol}_{k+1}  \ \Delta y^{fol}_{k \to k+1} \  v^{fol}_{y,k+1}  \ a^{fol}_{y,k+1} \ ]
\end{equation}
where $\mathcal{I}_{i}$ and $\mathcal{O}_{i}$ are the input features at time step $k$ and the output state labels at time step $k+1$, respectively. The subscripts represent the time step, $x^{fol}, v^{fol}$ and $a^{fol}$ are the lateral position, longitudinal velocity and acceleration of the following vehicle, respectively. $d^{rel}$ is the distance between the predicted vehicle and its leading vehicle, $v^{lead}$ is the velocity of the leading vehicle, and $\Delta y^{fol}_{k \to k+1}$ is the advancing distance of the predicted vehicle in the time interval (0.1 second).

The features were extracted from the raw trajectory data to generate training samples according to (\ref{short_train2})-(\ref{short_train3}). To guarantee the validity of training samples, the chosen pairs of leading vehicle and following vehicle must be adjacent in the same lane and not involved in lane-change events. We used 1.2 million training samples with 28 Gaussian mixture components to train the GSM.

In order to demonstrate the advantages of the proposed learning-based model, we compared them with several widely used models in tracking problems: constant velocity model (CVM), constant acceleration model (CAM) and intelligent driver model (IDM). For IDM, the parameters provided in \cite{r21} were utilized. 
Since they are deterministic models, we added Gaussian-distributed noise to provide uncertainties.
We used very noisy measurements to simulate partial occlusion; while under complete occlusion, no measurement except the initial state was provided thus there was no measurement update. Tracking under complete occlusion can be regarded as making long-term predictions.
We used 1,000 particle hypotheses to approximate the state distribution every 0.1 second for each model. The particle hypotheses obtained by iteratively propagating 100 time steps (10 seconds) from the initial state for partial occlusion and complete occlusion are visualized in Fig. 2 and Fig. 3, respectively. The particles are denoted as blue dots and the groundtruth are denoted as yellow stars. The Mean Absolute Error (MAE) of tracking results for partial occlusion is provided in Fig. 4. 

Fig. 2 demonstrates that GSM and CVM can maintain accurate tracking both on position and velocity. But compared with CVM, the particle-approximated state distribution of GSM has smaller variance, which means the conditional output distribution obtained from GSM is more centralized around groundtruth. The reason is that the fitted GSM is more robust to measurements with large noise than other models thus reduces the impacts of measurement noise. 
In Fig. 2(c), there is a delay and deviation on velocity tracking especially from time step 30 to 50, which means CAM is not good at tracking abrupt velocity changes. This can also be illustrated by error fluctuations in Fig. 4. 
In Fig. 2(d), the IDM tends to behave more conservative than human drivers thus anticipates a sudden brake to a lower speed than the ground truth and loses track of the velocity. This reveals that the parameters in \cite{r21} need to be carefully finetuned to adapt to different conditions. 
As demonstrated in Fig. 4, employing our GSM can achieve both the smallest position and velocity tracking error. 
A complementary video showing the multi-target tracking results can be found online.
\footnote{The demonstration video can be found on \url{https://berkeley.box.com/v/vehicle-tracking-demo}.}

Fig. 3 shows that although no model can track the position and velocity precisely under complete occlusion, our GSM is able to anticipate a deceleration due to reducing distance between itself and the leading vehicle and accelerations because of increasing gaps, which has a similar trend to the ground truth. The algorithm can recover tracking quickly when the target reappears as well as provide a reference for planning and control.
\begin{figure}[!tbp]\label{MAE}
	\centering
	\epsfig{figure=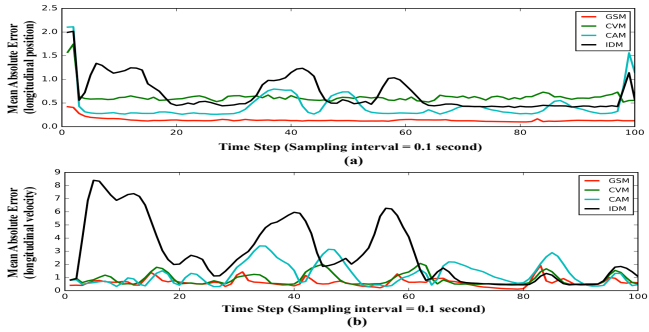,width=\columnwidth,height=4.8cm}
	\caption{Mean Absolute Error (MAE) of Longitudinal Position and Velocity (Lane Keeping)}
\end{figure}
\begin{figure}[htbp]\label{lane_change}
	\centering
	\epsfig{figure=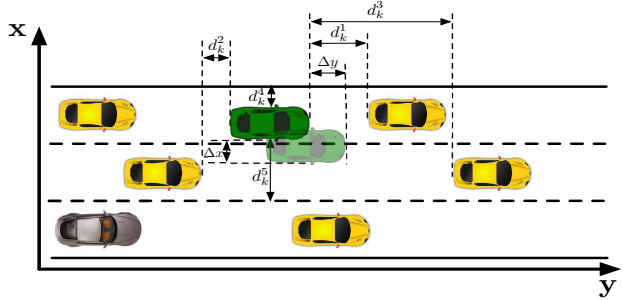,width=\columnwidth,height=4cm}
	\caption{Lane Change Behavior}
\end{figure}
\begin{figure*}[htbp]\label{lane_change_partial_occ}
	\centering
	\epsfig{figure=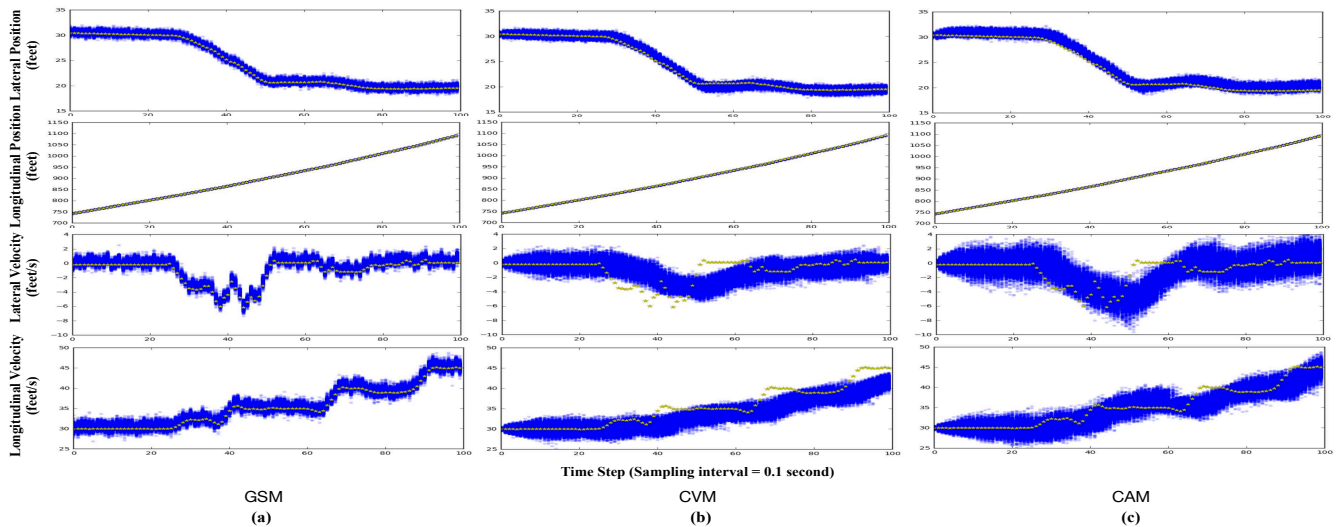,width=\textwidth,height=7cm}
	\caption{Lateral / Longitudinal Position and Velocity Tracking Results Comparison for Partial Occlusion (Lane Change)}
\end{figure*}
\begin{figure}[htbp]\label{MAE_lane_change}
	\centering
	\epsfig{figure=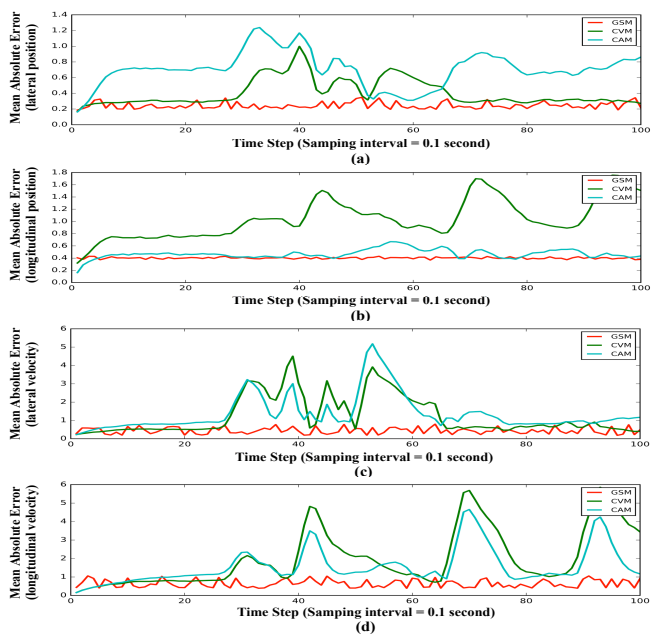,width=\columnwidth,height=8.5cm}
	\caption{Mean Absolute Error (MAE) of Lateral / Longitudinal Position and Velocity (Lane Change)}
\end{figure}
\subsection{Case 2: Lane Change}
In this case, the tracked vehicle is assumed to perform lane change behavior, which is illustrated in Fig. 5. Similar to the lane keeping case, we choose the green car as the predicted object. An assumption is applied that the yaw angles of vehicles always equal to zero since the ratio of lateral velocity to longitudinal velocity is very low in highway scenarios. 
The vehicle state includes lateral position, velocity, acceleration as well as longitudinal position, velocity, acceleration.
Since IDM is only able to model car-following behaviors, it is not used in this case.

When anticipating lane change motions, both the vehicles in the original lane and those in the target lane need to be considered. The notations and descriptions of input features and output state labels are listed in Table I. A total of 400 thousand training samples were selected from the trajectory segments covering 5 seconds before and after lane change moments.
The Gaussian mixture distribution was fitted with 60 Gaussian mixture components. 
Since the vehicles performing lane change behaviors is unlikely to be completely occluded for a long period, we only discuss partial occlusion instances.
\begin{table}[!tbp]
	\caption{Input Features and Output State Labels (Lane Change)}
	\label{tab:feature}
	\begin{center}
		\begin{tabular}{m{0.6cm} m{0.8cm} p{6cm}}
			\toprule
			\midrule
			&  \textbf{Notations} & \textbf{Descriptions} \\ 
			\midrule
			\multirow{14}*{\shortstack[lb]{Input \\ features}}  
			& $x_k$ & Lateral position \\ 
			& $y_k$ & Longitudinal position \\
			& $\dot{x}_k$ & Lateral velocity \\
			& $\dot{y}_k$ & Longitudinal velocity \\
			& $\ddot{x}_k$ & Lateral acceleration \\
			& $\ddot{y}_k$ & Longitudinal acceleration \\
			& $d^1_k$ & Distance to leading vehicle in the same lane \\
			& $d^2_k$ & Distance to following vehicle in the target lane \\
			& $d^3_k$ & Distance to leading vehicle in the target lane \\
			& $d^4_k$ & Distance to the left line of its original lane \\
			& $d^5_k$ & Distance to the right line of its target lane \\
			& $v^{rel}_{1,k}$ & Relative velocity to leading vehicle in the same lane \\
			& $v^{rel}_{2,k}$ & Relative velocity to following vehicle in target lane \\
			& $v^{rel}_{3,k}$ & Relative velocity to leading vehicle in the target lane \\
			\midrule
			\multirow{6}{*}[0cm]{\shortstack[lb]{Output \\ state \\ labels}} 
			& $\Delta x_{k : k+1}$ & Traveled lateral distance \\ 
			& $\Delta y_{k : k+1}$ & Traveled longitudinal distance \\
			& $\dot{x}_{k+1}$ & Lateral velocity at next time step \\
			& $\dot{y}_{k+1}$ & Longitudinal velocity at next time step\\
			& $\ddot{x}_{k+1}$ & Lateral acceleration at next time step \\
			& $\ddot{y}_{k+1}$ & Longitudinal acceleration at next time step \\
			\midrule
			\bottomrule
		\end{tabular}
	\end{center}
\end{table}
We compared particle distributions of our GSM, CVM and CAM in Fig. 6. It is illustrated that despite all three models can achieve acceptable position tracking accuracy, the GSM has greater advantages over CVM and CAM on velocity tracking, especially when immediate acceleration and deceleration happen. The MAE of lateral / longitudinal positions and velocities can be found in Fig. 7, in which the tracking errors of CVM and CAM have large fluctuations around turning points while GSM is able to maintain relatively stable performance.
A plausible reason is that given different input features, the conditional output distributions of GMM have different peaks which are able to give specified outputs for certain conditions.

\section{Conclusion and Future Work}
In this paper, a generic vehicle tracking framework based on modified mixture particle filter was proposed, where a learning-based behavioral model is incorporated as the system dynamics model in prior update. The proposed framework can not only adaptively set the number of tracking targets and allocate particles via updating the mixture representation at each time step, but also keep accurate tracking during occlusion periods by making short-term and long-term predictions. Two case studies including lane keeping and lane change scenarios were conducted to test its validation and efficiency, where a Gaussian mixture model was employed as an example of the learning-based model without loss of generality. 
For future work, the proposed tracking framework will be applied to more complicated scenarios such as intersections and round abouts. More advanced learning-based models such as deep neural network will be employed.







\end{document}